%% file: neurips_2025.tex
\documentclass{article}

\usepackage{natbib}
 \usepackage[preprint]{neurips_2025}
 \usepackage{graphicx}
 \usepackage{amsmath}
 \usepackage{graphicx}
\usepackage{booktabs}
\usepackage{multirow}
\usepackage{tabularx}
\usepackage{array}

\newlength\ImgW
\newcolumntype{C}{>{\centering\arraybackslash}m{\ImgW}}
\newcolumntype{L}{>{\raggedright\arraybackslash}m{0.18\linewidth}}

\newcommand{\img}[1]{\includegraphics[width=\linewidth]{#1}}
\newcommand{\nowrap}[1]{\mbox{\strut #1}}

\newcommand{\beforecells}[1]{%
  \img{Images/#1/org/osm_tile_default.png} &
  \img{Images/#1/org/specific_mask.png} &
  \img{Images/#1/org/generated.png}
}
\newcommand{\aftercells}[1]{%
  \img{Images/#1/edited/osm_tile_default.png} &
  \img{Images/#1/edited/specific_mask.png} &
  \img{Images/#1/edited/generated.png}
}
\newcommand{\tabrow}[2]{#1 & \beforecells{#2} & \aftercells{#2}\\}

\usepackage{graphicx,booktabs,tabularx,array}
\newcolumntype{Y}{>{\centering\arraybackslash}X}  % centered stretch columns
\newlength{\thumbh}
\newcommand{\cellimg}[1]{\includegraphics[height=\thumbh,keepaspectratio]{#1}}
\newcommand{\seasonrow}[1]{%
  \cellimg{Images/Seasonal/#1/OSM.png} &
  \cellimg{Images/Seasonal/#1/Specific_Mask.png} &
  \cellimg{Images/Seasonal/#1/Winter.png} &
  \cellimg{Images/Seasonal/#1/Summer.png}\\
}

\newcommand{\genrow}[1]{%
  \cellimg{Images/Generated/#1/OSM.png} &
  \cellimg{Images/Generated/#1/Specific.png} &
  \cellimg{Images/Generated/#1/Generated_1.png} &
  \cellimg{Images/Generated/#1/Generated_2.png} &
  \cellimg{Images/Generated/#1/Generated_3.png} &
  \cellimg{Images/Generated/#1/GT.png} \\
}

\workshoptitle{UrbanAI 2025}

\usepackage[utf8]{inputenc} % allow utf-8 input
\usepackage[T1]{fontenc}    % use 8-bit T1 fonts
\usepackage{hyperref}       % hyperlinks
\usepackage{url}            % simple URL typesetting
\usepackage{booktabs}       % professional-quality tables
\usepackage{amsfonts}       % blackboard math symbols
\usepackage{nicefrac}       % compact symbols for 1/2, etc.
\usepackage{microtype}      % microtypography
\usepackage{xcolor}         % colors

\title{OSMGen: Highly Controllable Satellite Image Synthesis using OpenStreetMap Data}

\author{%
 Amir Ziashahabi\footnotemark[1] \quad Narges Ghasemi\footnotemark[1] \quad Sajjad Shahabi \\
{\bfseries John Krumm \quad Salman Avestimehr \quad Cyrus Shahabi}\\
University of Southern California\\
\texttt{\{ziashaha, nghasemi, sajjadsh, jkrumm, salman, shahabi\}@usc.edu}
}

\begin{document}

\begingroup
\renewcommand\thefootnote{\fnsymbol{footnote}}
\footnotetext[1]{These authors contributed equally.}
\endgroup

% \begingroup
% \renewcommand\thefootnote{\fnsymbol{footnote}}
% \footnotetext[1]{These authors contributed equally.}
% \endgroup

\maketitle

\begin{abstract}
\input{Sections/Abstract}
\end{abstract}

\section{Introduction}

\input{Sections/Introduction}
\\
% \noindent\textbf{Related work.}
% \input{Sections/RelatedWork}

\section{Methodology}

\input{Sections/Background}
\input{Sections/Method}

\section{Experiments}
\input{Sections/Experiments}

% \section{Discussion and Conclusion}
\input{Sections/Conclusion}

\section{Acknowledgments}
\input{Sections/Acknowledgments}

% \section*{References}
 \bibliography{references}

%%%%%%%%%%%%%%%%%%%%%%%%%%%%%%%%%%%%%%%%%%%%%%%%%%%%%%%%%%%%
\newpage
\appendix
% \section*{Appendix}
\section{Technical Background}
\label{app:Background}

\input{Sections/Appendix/Background}

\section{Detailed Methodology}
\label{app:Method}
\input{Sections/Appendix/Method}

\section{Problem Context and Literature Review}
\label{app:problem_context}

\subsection{Limitations of Existing Satellite Image Generation Methods}
\label{app:limitations}
\input{Sections/Appendix/limitations}

\subsection{Advantages of Conditioning on OSM JSON}
\label{app:advantages}

\input{Sections/Appendix/advantage}

\section{Detailed Related Work}
\label{app:related_work}
\input{Sections/Appendix/related_work}

 % \section{Supplemental Figures}

 \section{Seasonal Variation}\label{app:seasonal}
 
\begin{figure}[!htbp]
\centering
\setlength{\tabcolsep}{6pt}
\renewcommand{\arraystretch}{1.0}
\scriptsize

\begin{tabularx}{0.9\linewidth}{*{4}{Y}}
\toprule
\textbf{Mask:\,General} & \textbf{Mask:\,Specific} & \textbf{Winter} & \textbf{Summer}\\
\midrule
\seasonrow{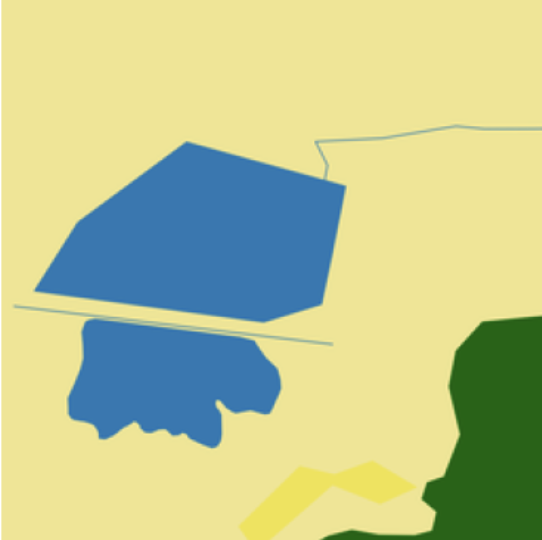}
\seasonrow{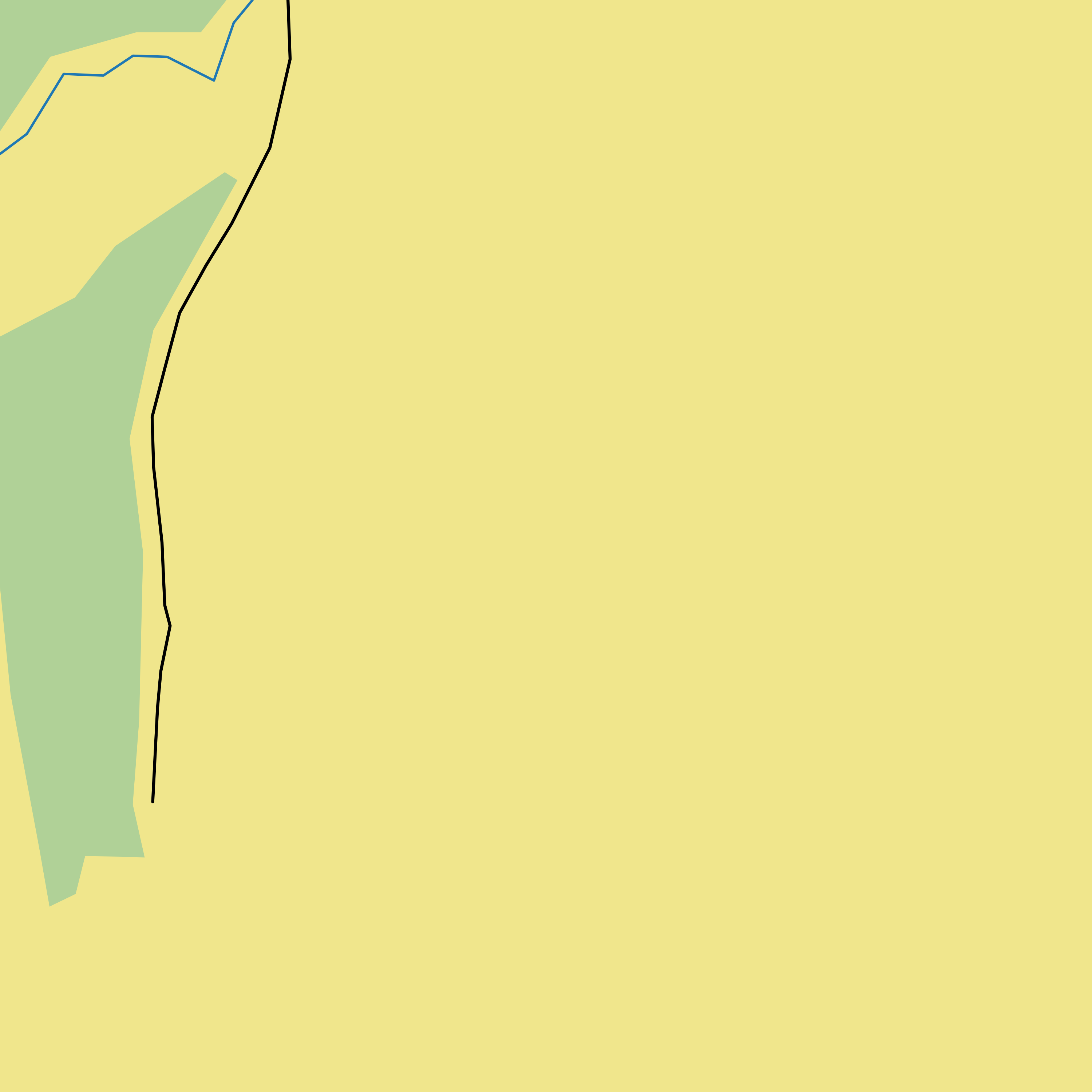}
\seasonrow{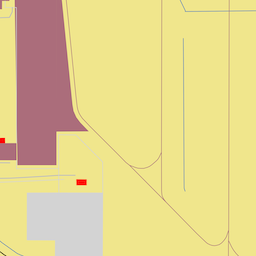}
\seasonrow{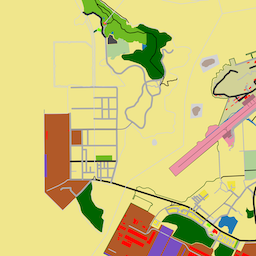}
\bottomrule
\end{tabularx}

\caption{Seasonal conditioning: for fixed masks, varying the date input produces distinct winter and summer images.}
\label{fig:seasonal}
\end{figure}

 To isolate the effect of temporal conditioning, we fix the semantic masks and text description for a single location and vary only the date input. This cleanly changes season‐specific appearance (e.g., vegetation density, color palette, and lighting) while leaving geometry unchanged. Sample images are provided in Figure~\ref{fig:seasonal}.
 % \begin{figure}[h!] \centering  \includegraphics[width=0.7\textwidth,keepaspectratio]{Images/season.png} \caption{Seasonal conditioning: for fixed masks, varying the date input produces distinct winter and summer images.} \label{fig:seasonal_appendix} \end{figure}

 % --- Figure ---

%%%%%%%%%%%%%%%%%%%%%%%%%%%%%%%%%%%%%%%%%%%%%%%%%%%%%%%%%%%%

% \newpage

\end{document}

%% file: Sections/Abstract.tex
% Effective urban planning and management depend critically on accurate and timely geospatial data. A significant barrier to automating urban monitoring, however, is the scarcity of large-scale, curated datasets capturing specific, isolated changes. To address this data bottleneck, we introduce a generative model that synthesizes high-fidelity satellite imagery directly from OpenStreetMap (OSM) data. Our approach conditions the image synthesis on a rich set of multi-modal inputs, allowing for fine-grained control over the generated scene. The key innovation is the ability to generate consistent "before and after" image pairs, where only user-defined changes in the OSM data are reflected, while all other scene elements remain identical. This capability enables the targeted synthesis of underrepresented urban features and infrequent change events, directly addressing the critical challenges of data scarcity and class imbalance for training robust detection models. Consequently, our method provides two primary benefits: it allows for the generation of tailored training data to monitor critical urban dynamics, such as unauthorized construction or land-use conversion; and it serves as a powerful simulation tool for planners to visualize and assess the impact of proposed interventions, thereby facilitating more proactive and data-driven urban development.

Accurate and up-to-date geospatial data are essential for urban planning, infrastructure monitoring, and environmental management. Yet, automating urban monitoring remains difficult because curated datasets of specific urban features and their changes are scarce. We introduce OSMGen, a generative framework that creates realistic satellite imagery directly from raw OpenStreetMap (OSM) data. Unlike prior work that relies on raster tiles, OSMGen uses the full richness of OSM JSON, including vector geometries, semantic tags, location, and time, giving fine-grained control over how scenes are generated. A central feature of the framework is the ability to produce consistent before–after image pairs: user edits to OSM inputs translate into targeted visual changes, while the rest of the scene is preserved. This makes it possible to generate training data that addresses scarcity and class imbalance, and to give planners a simple way to preview proposed interventions by editing map data. More broadly, OSMGen produces paired (JSON, image) data for both static and changed states, paving the way toward a closed-loop system where satellite imagery can automatically drive structured OSM updates. 
Source code is available at \url{https://github.com/amir-zsh/OSMGen}.

%% file: Sections/Introduction.tex
Urban planning can greatly benefit from accurate and timely geospatial data \cite{reynard2018five,lee2015geospatial}. A readily available source is OpenStreetMap (OSM) \cite{OpenStreetMap}, a collaborative project that offers more than just a visual map: it provides a detailed, structured JSON format containing rich information lost in simple image tiles, such as precise vector geometries and semantic tags for every feature. While this detailed data structure is ideal for conditioning a generative process, it has been largely underexplored for this purpose, partly due to the complexity of the JSON files.

In this work, we leverage the full depth of OSM data for image generation. The core of our contribution is a novel generative model that synthesizes high-fidelity satellite imagery by conditioning on the structured information within OSM JSON. In contrast to methods that use rendered map images, our approach utilizes a richer set of inputs including feature tags, location, and date, enabling highly controllable and precise synthesis. We also introduce a method to leverage this model for controlled scene manipulation: by editing the OSM-derived inputs, we can synthesize a corresponding "after" image that is perfectly co-registered to its "before" state, thereby isolating the visual impact of a single, defined change.

Our approach enables two powerful applications. First, it can generate vast, pixel-perfect labeled datasets to address data scarcity in geospatial AI, improving downstream models for tasks like building footprint segmentation \cite{li2022review} and land-use classification \cite{zhu2017deep}. Second, it serves as a dynamic simulation tool, allowing urban planners to visualize the impact of proposed developments, such as new parks or infrastructure, by simply editing the JSON map data, thus supporting data-driven decision-making.

Crucially, our framework is unique in its ability to generate complete, corresponding pairs of (JSON, image) data for both before and after states. This provides the exact data required to train the next generation of cartographic models that can truly "close the loop": detecting changes in new satellite imagery to suggest automated, structured updates to OSM JSON. This capability promises to significantly reduce the manual effort needed to keep the world's map current and accurate \cite{bastani2021updating}.

\paragraph{Background.} Denoising diffusion models synthesize images by learning to reverse a fixed process of gradually adding noise \cite{ho2020denoising, sohl2015deep}. By training a neural network to perform this reverse denoising operation, the model can generate new, high-fidelity images starting from pure noise. To enable precise, training-free image editing, we leverage Denoising Diffusion Implicit Models (DDIM) \cite{song2020denoising}, which introduce a deterministic variant of the diffusion process. This determinism is crucial because it allows the process to be inverted. Given a real image, DDIM inversion can trace the denoising path backward to find the unique latent code that generates it. This inversion capability is the key mechanism that enables high-fidelity, targeted modifications. The technical details of these processes are detailed further in Appendix~\ref{app:Background}.

%% file: Sections/Method.tex
This work presents an end-to-end pipeline for generating satellite imagery conditioned on rich information derived from raw OSM JSON data. Our approach addresses key shortcomings of methods that rely on simpler inputs such as raster tiles or bounding boxes, which lack the precise geometries and detailed tag-level semantics available in the source OSM data \cite{zang2025changediff,10887344}; see Appendix~\ref{app:limitations} for a detailed analysis. By leveraging the source JSON, we enable fine-grained, controllable, and spatially accurate synthesis. Please refer to Appendix~\ref{app:related_work} for a detailed literature review on image generation, satellite image generation, and image editing.

\begin{figure}[!t]
  \centering
  \includegraphics[width=\textwidth]{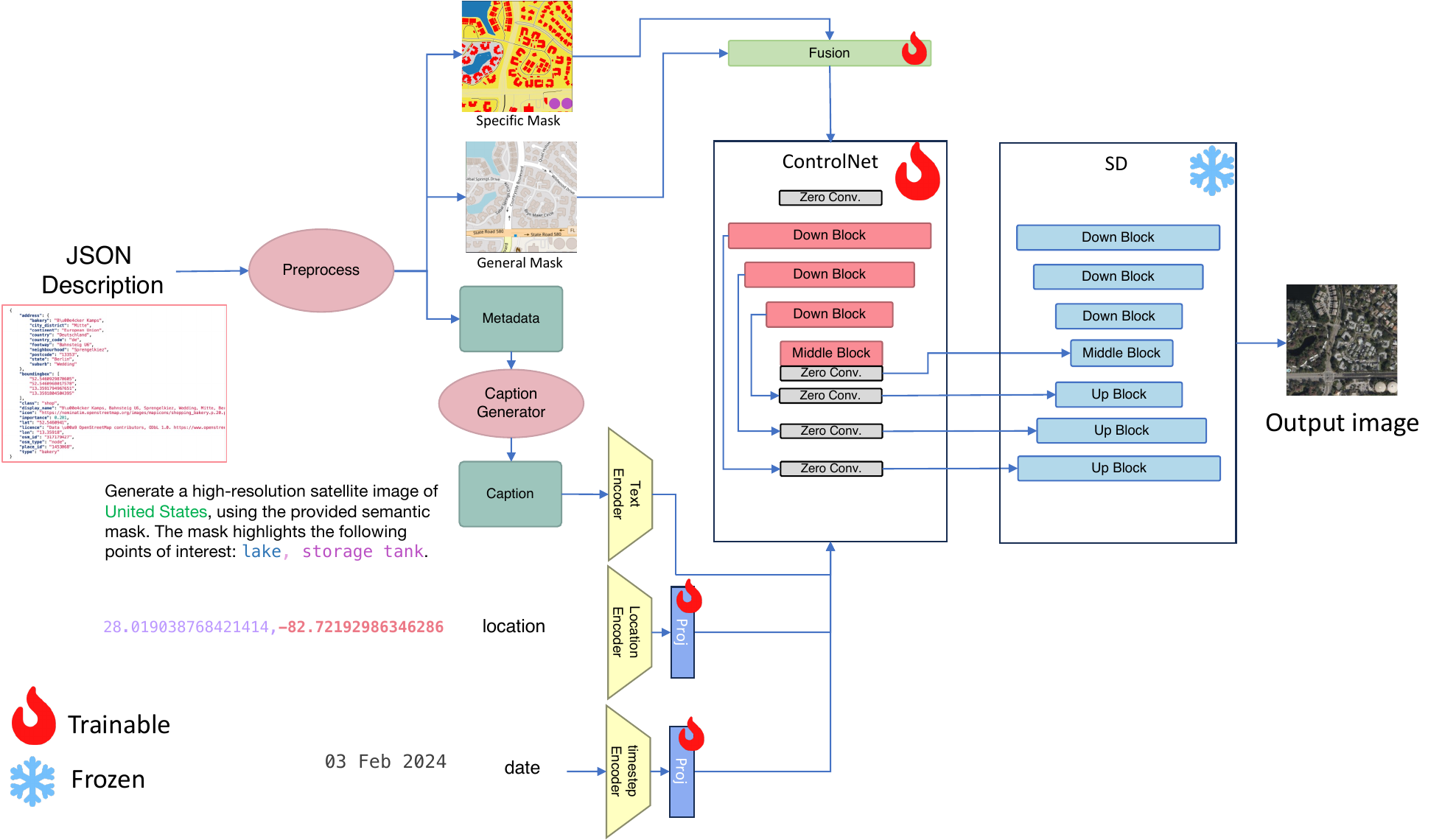}
  \caption{Overview of our ControlNet pipeline. Semantic masks are fed into ControlNet to generate control feature maps that are added into the U-Net; spatial and temporal embeddings are summed into the timestep embedding; the text prompt is injected via cross-attention.}
  \label{fig:pipeline}
\end{figure}

\paragraph{Data Collection and Preprocessing.}
To ensure broad geographic coverage, we sample approximately 20{,}000 points from the Functional Map of the World (FMoW) benchmark \cite{christie2018functional}, spanning urban centers, suburbs, and rural areas. For each point, we fix a zoom level \(z\) in advance (typically choosing \(z=18\) to capture fine-grained structural details and \(z=15\) for wider contextual views) and compute the exact \(256\times 256\)-pixel tile bounds around the center latitude and longitude via standard Web Mercator tile formulas \cite{OSMSlippyMap}. Using this bounding box and zoom, we retrieve a \(256\times 256\) satellite image tile and its corresponding raw OSM JSON for each data point. From the JSON, we extract a multimodal set of conditions designed to provide comprehensive guidance to the generative model. The primary conditions are two segmentation masks derived from the raw vector geometries: (1) the \textbf{general mask}, which groups features into a small number of broad categories such as roads, water bodies, vegetation, buildings, and other primary surface types, capturing high-level concepts; and (2) the \textbf{specific mask}, which assigns each fine-grained point-of-interest (POI) subtype (e.g., lakes, rivers, storage tanks, solar farms) its own mask color so the model can learn the nuances of each type. To capture spatiotemporal context, we encode the tile’s geographic coordinates using SatCLIP \cite{klemmer2025satclip} and its capture date using Date2Vec \cite{date2vec}. Finally, we generate a textual summary of the tile’s most salient categories and encode it via a frozen CLIP text encoder \cite{radford2021learning} to provide high-level semantic guidance (see Appendix~\ref{app:encoders} for details).

\paragraph{Generation Framework.}
As illustrated in Figure~\ref{fig:pipeline}, our framework augments a frozen Stable Diffusion U-Net~\cite{rombach2022high} with a trainable ControlNet branch~\cite{zhang2023adding}. The general and specific masks are fused via a convolutional layer and provided to ControlNet to enforce geometric fidelity. Spatial and temporal embeddings are each passed through a linear projection and then added to the diffusion timestep embedding, while the text embedding is injected through cross-attention. We train the ControlNet component, the mask-fusion layer, and the linear projections for spatial and temporal conditioning, using the standard diffusion loss:
\[
\mathcal{L}_{\mathrm{diff}}
= \mathbb{E}_{x_{0},\,t,\,\epsilon}
\Bigl\lVert
\epsilon \;-\; \epsilon_{\theta}\bigl(x_{t},\,t \mid M,\mathbf{e}_{\mathrm{loc}},\mathbf{e}_{\mathrm{time}},\mathbf{e}_{\mathrm{text}}\bigr)
\Bigr\rVert_{2}^{2}.
\]
Here, the network $\epsilon_{\theta}$ is trained to predict the ground-truth noise $\epsilon$ from the noisy image latent $x_t$ at timestep $t$, given the set of conditions: the fused mask $M$ and the location, time, and text embeddings ($\mathbf{e}_{\mathrm{loc}}, \mathbf{e}_{\mathrm{time}}, \mathbf{e}_{\mathrm{text}}$).

\paragraph{Controlled Change Generation.}
To create consistent before/after image pairs, we use DDIM inversion \cite{song2020denoising}. This choice is driven by three factors. First, cross-attention–based editing methods are unsuitable here because they do not account for our nontextual conditions (masks, spatial, and temporal information) \cite{hertz2022prompt}. Second, DDIM inversion is straightforward to implement and agnostic to the model’s specific architecture. Third, strong spatial conditioning from the masks allows us to reduce the classifier-free guidance (CFG) scale, mitigating a known limitation of DDIM inversion, instability at high CFG scales \cite{mokady2023null,huberman2024edit}, and yielding high-fidelity results. Please refer to Appendix \ref{app:ddim_inversion} for details.

%% file: Sections/Experiments.tex
% Given the absence of an off-the-shelf benchmark for OSM-conditioned satellite synthesis and the substantial compute resources required for large-scale quantitative metrics, we focus here on a qualitative assessment of our model’s fidelity to both geometry and semantics.
% \subsection{Experimental Setup and Qualitative Results} 

\begin{figure}[!t]
\centering

\begin{minipage}{0.8\linewidth} % was 1.1; smaller container too (optional)
  \setlength{\tabcolsep}{6pt}
  \renewcommand{\arraystretch}{1.0}
  \scriptsize

  % SHRINK IMAGES: lower the per-cell image height (was 22mm)
  \setlength{\thumbh}{16mm}   % try 16mm, or 14mm if you want even smaller
  \setlength{\ImgW}{0.13\linewidth} % keep your column width ratio

  \begin{tabular}{*{6}{C}} % uses C := m{\ImgW}
    
    \text{Mask: General} & \text{Mask: Specific} &
    \text{Generated \#1} & \text{Generated \#2} &
    \text{Generated \#3} & \text{Ground Truth} \\
    \midrule
    \genrow{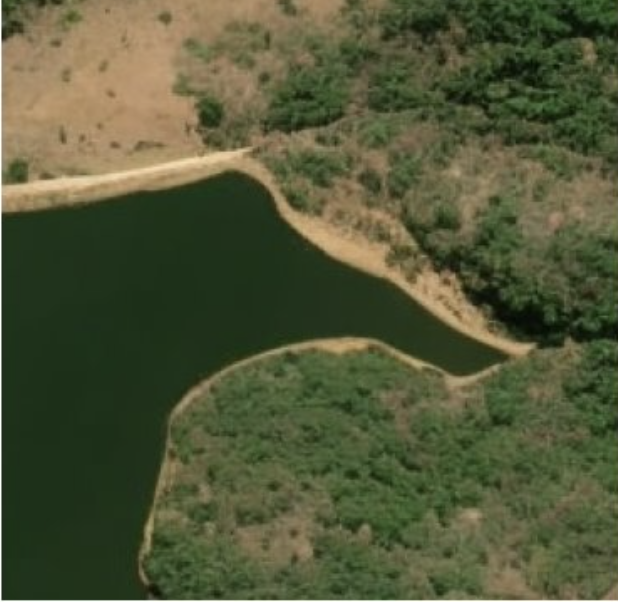}
    \genrow{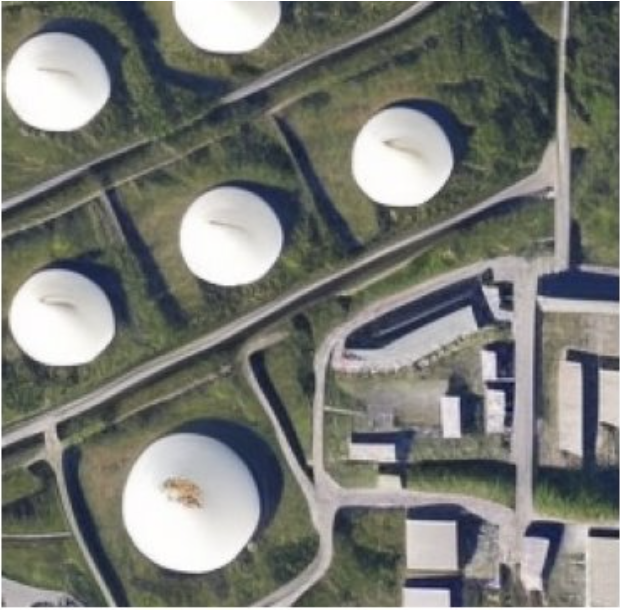}
    \genrow{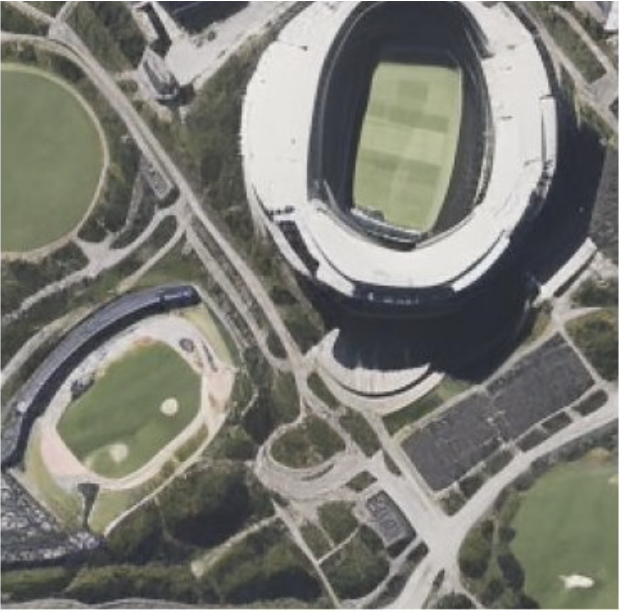}
  \end{tabular}
\end{minipage}

\caption{Qualitative evaluation on held‐out FMoW locations. This layout highlights both the model’s ability to reproduce large‐scale structure and to capture fine‐grained POI details in context.}
% % Each row shows, from left to right: the input general mask, the input specific mask, three independently generated satellite images, and the ground truth.}
\label{fig:qualitative}
\end{figure}

\paragraph{Experimental Setup} We trained the model for 500 epochs using a batch size of 2048. Our evaluation uses samples from a held-out test set of approximately 2,000 locations from our FMoW-derived dataset. For each location, we generate a 256x256 pixel tile using the full multi-modal conditioning framework described previously. All synthesis operations were performed on a single NVIDIA A100 GPU.
\paragraph{Qualitative Results.} Figure~\ref{fig:qualitative} shows representative outputs, demonstrating that the model (i) accurately reproduces road networks and building footprints from the general mask and (ii) renders rare POI classes (e.g., stadiums, storage tanks) with correct shapes and context from the specific mask.
% \subsection{Analysis of Conditioning and Change Generation}
A brief analysis of seasonality under temporal conditioning is deferred to Appendix~\ref{app:seasonal}.
\textbf{Consistency via DDIM Inversion:} We apply DDIM inversion and re-denoising with an edited mask to generate “after” images in which regions outside the edited area remain consistent with the “before” state. Figure~\ref{fig:ddim} demonstrates samples produced using this method for edits that add, remove, or modify elements. The pipeline produces consistent pairs without introducing artifacts outside the intended changes.

% \begin{figure}[h!] \centering \includegraphics[width=\textwidth]{Images/Examples.jpg} \caption{Qualitative evaluation on held‐out FMoW locations. This layout highlights both the model’s ability to reproduce large‐scale structure and to capture fine‐grained POI details in context.
% % Each row shows, from left to right: the input general mask, the input specific mask, three independently generated satellite images, and the ground truth.
% } \label{fig:qualitative} \end{figure} 

% \begin{figure}[h!] \centering  \includegraphics[width=0.6\textwidth,keepaspectratio]{Images/Change.pdf} \caption{Change synthesis via DDIM inversion. The pipeline produces consistent “before”/“after” pairs.} \label{fig:ddim} \end{figure} 

% \paragraph{Discussion.} qualitative results demonstrate that the proposed ControlNet-augmented diffusion pipeline produces satellite images that faithfully adhere to both the large-scale layout and fine-grained semantics specified by raw OSM JSON. By incorporating DDIM inversion, we enable deterministic reconstruction and controlled variation of individual conditioning signals. This inversion-driven editing workflow confirms that our model achieves \textbf{geometric consistency}, where the overall scene structure remains anchored even when specific features are altered; \textbf{semantic flexibility}, allowing for precise changes like adding or removing POIs without perturbing unrelated elements; and \textbf{temporal adaptability}, yielding seasonally plausible variations by modifying only the time input. 

% --- Figure ---
\begin{figure}[!t]
\centering
\setlength{\tabcolsep}{4pt}
\renewcommand{\arraystretch}{1.0}
\scriptsize

\begin{tabular}{L *{6}{C}}
 & \multicolumn{3}{c}{Before} & \multicolumn{3}{c}{After} \\
\cmidrule(lr){2-4}\cmidrule(lr){5-7}
\textbf{Edit} &
\nowrap{Mask: General} & \nowrap{Mask: Specific} & \nowrap{Generated} &
\nowrap{Mask: General} & \nowrap{Mask: Specific} & \nowrap{Generated} \\
\midrule
\tabrow{\textbf{ADD} a stadium}{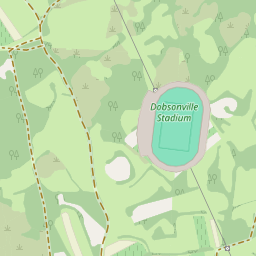}
\tabrow{\textbf{ADD} a building}{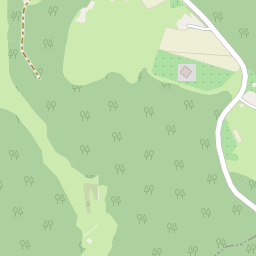}
\tabrow{\textbf{Remove} some buildings}{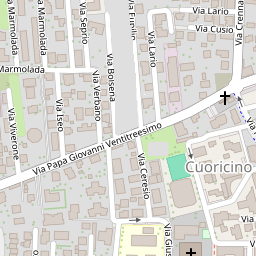}
\tabrow{\textbf{Remove} some storage tanks}{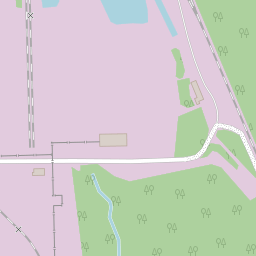}
\tabrow{\textbf{Change} lake to grass }{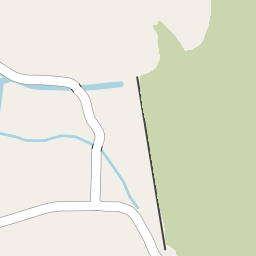}
\tabrow{\textbf{Change} crop field to solar farm}{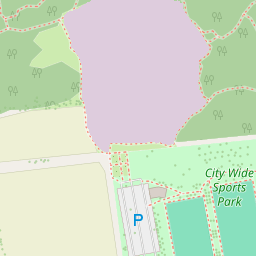}
\end{tabular}

\caption{Editing via DDIM inversion. Edits are applied locally while preserving consistency outside the modified region.}

\label{fig:edit-grid}
\end{figure}

%% file: Sections/Conclusion.tex
\section{Conclusion} We have presented a novel approach for high-fidelity satellite image synthesis conditioned on OSM JSON. This framework achieves both structural fidelity and semantic richness, opening new avenues for interactive geospatial content creation and data augmentation. This framework paves the way for generating large labeled datasets, including static imagery and co-registered before/after pairs, to be used in downstream tasks, such as segmentation, change detection, and automated proposals for OSM JSON updates based on satelite image changes.

%% file: Sections/Acknowledgments.tex
This research has been funded in part by NSF grants IIS-2128661 and 1956435, and NIH grant
5R01LM014026. Opinions, findings, conclusions, or recommendations
expressed are those of the author(s) and do not necessarily reflect the views of any sponsors, such as NSF or NIH.

%% file: Sections/Appendix/Background.tex
\subsection{DDPM}

\paragraph{Diffusion Process} \cite{ho2020denoising, sohl2015deep}
consists of two processes: a tractable forward process that gradually adds Gaussian noise to data, and a learned reverse process that recovers clean data from noisy inputs. Specifically, the forward process is a Markov chain defined as
\begin{align*}
q(x_{t} \mid x_{t-1}) = \mathcal{N}\!\bigl(x_{t}; \sqrt{1 - \beta_{t}}\,x_{t-1}, \,\beta_{t}\mathbb{I}\bigr),
\end{align*}
with a variance schedule $\beta_{1:T}$. This yields the joint distribution
\begin{align*}
q(x_{1:T} \mid x_0) = \prod_{t=1}^{T} q(x_t \mid x_{t-1}),
\end{align*}
and a tractable marginal for $x_t$ given $x_0$:
\[
q(x_t \mid x_0) = \mathcal{N}\!\bigl(x_t; \sqrt{\bar{\alpha}_t}\,x_0, \,(1 - \bar{\alpha}_t)\mathbb{I}\bigr),
\]
where $\bar{\alpha}_t = \prod_{s=1}^t (1 - \beta_s)$. While the forward process is fixed, the true reverse process $q(x_{t-1}\!\mid\!x_t)$, which is required for generation, is intractable to compute directly; in DDPM (\emph{Denoising Diffusion Probabilistic Models}; \cite{ho2020denoising}) this reverse step is approximated by a parameterized model $p_\theta$, typically implemented as a neural network:
\[
p_\theta(x_{t-1} \mid x_t) = \mathcal{N}\!\bigl(x_{t-1} \mid \mu_\theta(x_t, t), \,\Sigma_\theta(x_t, t)\bigr).
\]
When side information $c$ (e.g., text, mask, layout) is available, the reverse model can be conditioned on $c$:
\[
p_\theta(x_{t-1} \mid x_t, c) = \mathcal{N}\!\bigl(x_{t-1} \mid \mu_\theta(x_t, t, c), \,\Sigma_\theta(x_t, t, c)\bigr).
\]
% In the common noise-prediction parameterization we write $\epsilon_\theta(x_t,t,c)$ and the corresponding clean estimate
% $\hat x^{(t)}_0(x_t,c) = \frac{x_t-\sqrt{1-\alpha_t}\,\epsilon_\theta(x_t,t,c)}{\sqrt{\alpha_t}}$.

\subsection{DDIM Inversion}\label{app:ddim_inversion}

Denoising Diffusion Implicit Models (DDIM) \cite{song2020denoising} relax the Markovian assumption of DDPM by introducing a family of non-Markovian transitions. Using the reparameterization trick, the reverse step can be written as
\[
\boldsymbol{x}_{t-1}
= \sqrt{\alpha_{t-1}} \underbrace{\left(\frac{\boldsymbol{x}_t-\sqrt{1-\alpha_t}\, \epsilon_\theta^{(t)}\!\left(\boldsymbol{x}_t,\!c\right)}{\sqrt{\alpha_t}}\right)}_{\text{predicted } \boldsymbol{x}_0}
+ \underbrace{\sqrt{1-\alpha_{t-1}-\sigma_t^2}\; \epsilon_\theta^{(t)}\!\left(\boldsymbol{x}_t,\!c\right)}_{\text{direction toward } \boldsymbol{x}_t}
+ \underbrace{\sigma_t \,\epsilon_t}_{\text{random noise}},
\]
where $\epsilon_t\sim\mathcal{N}(0,\mathbb{I})$ is independent noise at step $t$. Setting
\[
\sigma_t=\sqrt{\frac{1-\alpha_{t-1}}{1-\alpha_t}} \sqrt{1-\frac{\alpha_t}{\alpha_{t-1}}}
\]
recovers the stochastic DDPM sampler (Markovian), and setting $\sigma_t = 0$ yields a deterministic update, i.e., the DDIM transition. DDIM allows skipping intermediate noise levels for faster sampling and, due to its determinism, enables precise inversion for image editing.

\textbf{Inversion.} Choose an inversion depth $t^\star\!\in\!\{0,\ldots,T\}$ that sets the \emph{edit strength}. Starting from an observed image $x_{\text{obs}}$ and a \emph{reference} condition $c_{\mathrm{ref}}$, apply the deterministic DDIM forward (noise-adding) updates only up to $t^\star$:
\[
x_{t+1} = \sqrt{\alpha_{t+1}}\!\Bigl(\tfrac{x_t - \sqrt{1 - \alpha_t}\,\epsilon_\theta^{(t)}(x_t,c_{\mathrm{ref}})}{\sqrt{\alpha_t}}\Bigr)
+ \sqrt{1 - \alpha_{t+1}}\,\epsilon_\theta^{(t)}(x_t,c_{\mathrm{ref}}), \qquad t=0,\ldots,t^\star-1,
\]
initialized at $x_0=x_{\text{obs}}$. The latent $x_{t^\star}$ is the inversion endpoint: $t^\star{=}T$ gives full inversion; smaller $t^\star$ preserves more of $x_{\text{obs}}$ (weaker edits). With $\sigma_t{=}0$, running the matching DDIM denoiser from $t^\star$ back to $0$ under the same condition approximately reconstructs $x_{\text{obs}}$.

\textbf{Redenoising with new conditions.} To edit, re-denoise the inverted latents under a \emph{target} condition $c_{\mathrm{new}}$ using the DDIM update
\[
\tilde x_{t-1} \;=\; \sqrt{\alpha_{t-1}}\!\Bigl(\tfrac{x_t - \sqrt{1-\alpha_t}\,\epsilon_\theta^{(t)}(x_t,c_{\mathrm{new}})}{\sqrt{\alpha_t}}\Bigr)
+ \sqrt{1-\alpha_{t-1}}\,\epsilon_\theta^{(t)}(x_t,c_{\mathrm{new}}), \qquad t=t^\star,\ldots,1,
\]
initialized at the encoded $x_{t^\star}$ from the inversion pass. If $c_{\mathrm{new}}=c_{\mathrm{ref}}$ and the same $t^\star$ and schedule are used, the procedure approximately reconstructs $x_{\text{obs}}$; otherwise, it produces an edited output consistent with $c_{\mathrm{new}}$.

\textbf{Edit strength.} The single knob $t^\star$ implements the fidelity–edit trade-off introduced above: smaller $t^\star$ leads to higher fidelity / weaker edits; larger $t^\star$ (up to $T$) leads stronger edits. 

\begin{figure}[h!] \centering
\includegraphics[width=0.9\textwidth,keepaspectratio]{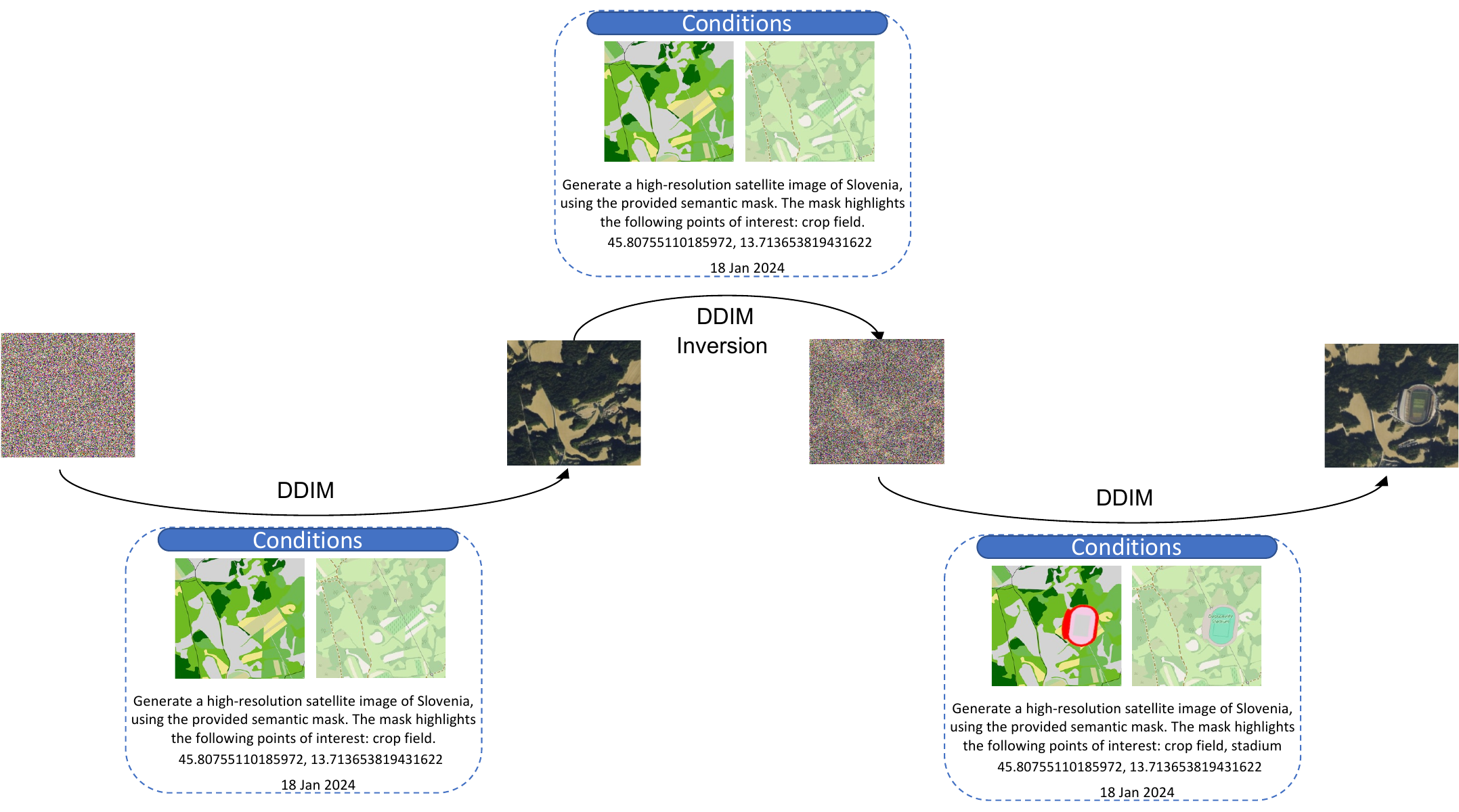}
\caption{Change synthesis via DDIM inversion.}
\label{fig:ddim}
\end{figure}

Figure~\ref{fig:ddim} illustrates this pipeline within our framework to produce consistent edits.

%% file: Sections/Appendix/Method.tex
\subsection{Component Encoders and Conditioning} \label{app:encoders} \paragraph{Mask Conditioning via ControlNet} We first combine the general and specific segmentation masks by stacking them into a multi‐channel tensor and passing this tensor through a small multilayer perceptron (MLP) that projects the concatenated mask channels down to the control‐image dimension required by ControlNet. The resulting fused mask embedding is then supplied to ControlNet’s image encoder. At each U-Net block, ControlNet produces control features from this embedding and adds them to the corresponding feature maps in the diffusion network, ensuring that the generated output strictly follows the prescribed geometries and class layout. \paragraph{Spatial Conditioning (Location Encoder)} We encode the tile center’s longitude–latitude pair \((\lambda,\phi)\) using SatCLIP \cite{klemmer2025satclip}. SatCLIP projects \(\lambda\) and \(\phi\) into a multi-scale sinusoidal basis and refines them via a two-layer MLP to produce a \(D\)-dimensional vector \(\mathbf{e}_{\mathrm{loc}}\). We then pass \(\mathbf{e}_{\mathrm{loc}}\) through a learnable linear projection before adding it to the diffusion timestep embedding at every denoising iteration, thereby injecting geographic context into the noise schedule.

\paragraph{Temporal conditioning (time encoder).}
To capture seasonal and illumination effects, we encode the capture timestamp using Date2Vec\cite{date2vec}, a pretrained, Time2Vec~\cite{kazemi2019time2vec}–inspired encoder. Given a 6-D timestamp vector (hh:mm:ss, yyyy-mm-dd), Date2Vec produces an embedding comprising a learned linear component and periodic (sinusoidal) components. As with the spatial pathway, we apply a learnable linear projection to this embedding before adding it to the model’s timestep embedding.

% \paragraph{Temporal Conditioning (Time Encoder)} To support seasonal and lighting variations, we apply Time2Vec \cite{kazemi2019time2vec} to the satellite image's capture date \(t\). Time2Vec outputs a vector \[ \mathbf{e}_{\mathrm{time}} = [\,\omega_{0} t + b_{0},\,\sin(\omega_{1} t + b_{1}),\dots,\sin(\omega_{K} t + b_{K})\,], \] combining a linear drift term with multiple learned sinusoids. Similar to the location encoding, \(\mathbf{e}_{\mathrm{time}}\) is summed into the timestep embedding, allowing the model to adapt synthesis to temporal context. 

\paragraph{Textual Conditioning (Prompt Encoder)} We generate a concise prompt for each tile, for example: \begin{quote} “Generate a high‐resolution satellite image in \emph{Country}, using semantic masks highlighting \emph{$POI_1$, $POI_2$, …}.” \end{quote} This prompt is tokenized and encoded by a frozen CLIP text encoder \cite{radford2021learning} into \(\mathbf{e}_{\mathrm{text}}\). In the main U-Net branch, \(\mathbf{e}_{\mathrm{text}}\) is injected via cross-attention, complementing the strict mask and embedding conditioning with flexible, human-readable guidance. 
% \subsection{Architectural Details} \label{app:architecture} We start from a pretrained Stable Diffusion model \cite{rombach2022high}, whose U-Net denoiser remains frozen during our training. Parallel to the main U-Net, we attach a ControlNet branch \cite{zhang2023adding} with identical down‐ and up‐sampling blocks but initialized with zero‐convolution (“zero conv.”) layers. This branch processes the concatenated general and specific masks \(M\) and produces “control features” that are added into the corresponding feature maps of the main U-Net, ensuring that the generated image strictly follows the prescribed geometry. 

%% file: Sections/Appendix/limitations.tex
Most prior work conditions satellite image synthesis on raster map tiles or on text prompts augmented with bounding‐box masks. In the raster‐tile approach \cite{zang2025changediff,zheng2024changen2}, rich vector data are flattened into a fixed palette of colors and symbols, erasing tag‐level distinctions among related feature types (for example, different road classes or categories of commercial establishments) and forcing generative models to treat semantically distinct entities as identical. Consequently, these methods cannot selectively render subtypes of interest at inference time.

When using text descriptions alongside simple rectangular masks\cite{10887344,tang2024crs}, each geographic feature is reduced to an axis‐aligned box that encloses its true shape. This approximation discards the precise polygonal outlines of buildings, the continuous centerlines of roads, and the irregular boundaries of land‐use areas, resulting in spatial misalignment in the synthesized imagery.

Moreover, reliance on a fixed iconography and color scheme limits extensibility. Introducing new feature classes or applying custom styling to existing ones requires manual tile redesign or retraining on freshly collected raster datasets, which undermines the goal of fine‐grained, user‐driven control. These inherent shortcomings motivate our shift to conditioning directly on raw OSM JSON, where exact geometries and explicit semantic tags remain fully available.

%% file: Sections/Appendix/advantage.tex
Directly using OSM JSON grants access to rich, structured metadata that raster tiles or bounding box approaches lack. Every feature in the JSON (nodes, ways, and relations) is annotated with comprehensive key–value tags (e.g.\ \texttt{amenity=restaurant}, \texttt{shop=bakery}), as well as hierarchical relations and auxiliary attributes. This level of detail enables the generator to distinguish and render closely related subclasses of features, rather than treating them as uniform pixels.

Moreover, OSM JSON encodes exact vector geometries such as polygonal footprints for buildings and land‐use areas, and linear centerlines for roads, which we convert into masks to preserve spatial fidelity. The JSON schema’s flexibility supports fine‐grained control over the conditioning inputs: users can select specific feature types, adjust per‐class palettes, or introduce and use various categories of entities. This deterministic mapping from vector layout to image conditioning ensures fully controllable and repeatable generation.

%% file: Sections/Appendix/related_work.tex
% Include a literature review of prior related work here.

\subsection{Image Generation}
Diffusion models \cite{sohl2015deep, ho2020denoising}  introduced a new powerful approach for image generation and have demonstrated superior performance over existing methods such as Generative Adversarial Networks (GANs) \cite{goodfellow2014generative} and Variational Auto Encoders (VAEs) \cite{kingma2013auto}, often yielding higher-fidelity images \cite{dhariwal2021diffusion}. 
Ho et al. \cite{ho2020denoising} put diffusion models on the map by demonstrating their capability in generating high-quality images. 
Latent Diffusion Models (LDM) \cite{rombach2022high} improved upon image generation using diffusion by employing the diffusion process in the latent space instead of the pixel space, resulting in reduced computational complexity and higher fidelity generated images. Furthermore, they used a text encoder alongside cross-attention to enable conditional generation of images based on input text. Stable Diffusion \cite{stabilityai2022stable} uses the same architecture as LDM and trains the model on a much larger dataset. Furthermore, they improve upon the text encoding process by using CLIP \cite{radford2021learning} as the text encoder. ControlNet \cite{zhang2023adding} utilizes the Stable Diffusion model and adds extra conditions to the generation process besides text. They enforce the extra conditioning by copying and freezing the pretrained base model. The copied version is trainable and used for learning new conditions, and the output of this model is added back to the frozen model using zero convolution. This enables the model for fine-grained, structure-preserving generation
guided by various conditioning inputs like edge maps, depth information, segmentation masks, or
human pose estimations.

\subsection{Satellite Image Generation}

Recent advances in layout-to-image generation have paved the way for controllable synthesis. In these works, input layouts, which are often represented as bounding boxes, segmentation masks, or other structured formats, serve as the primary condition for image generation. Several works have employed rectangular box layouts for controllable generation, including \cite{zheng2023layoutdiffusion, cheng2023layoutdiffuse, he2021context, zhao2020layout2image, qu2023layoutllm}. In the context of satellite image generation, one approach is to convert spatial relationship descriptions into structured layouts that direct image synthesis. For example, Lei et al.\cite{10887344} propose a two-stage framework that transforms spatial relationship descriptions into structured layouts and then synthesizes the final image using an enhanced diffusion model with positional prompts and layout attention. Although this method produces highly spatially accurate results, its reliance on fixed layouts and a limited set of classes restricts flexibility in diverse regions. CC-Diff \cite{zhang2024cc} enhances contextual coherence by integrating a dual resampler with foreground-aware attention to align the generated foreground with the background, yet it does not incorporate additional metadata that could improve output control.

Another approach is generating satellite images via image-to-image translation. Some works use GANs for this purpose, including CycleGAN~\cite{zhu2017unpaired} and pix2pix~\cite{isola2017image}. In particular, pix2pix employs a conditional adversarial framework with a U-Net generator and a PatchGAN discriminator to translate input images into outputs. Although this method can be applied to map-to-satellite conversion, it struggles to deliver high-quality results when the input maps are complex. Diffusion-based approaches have also been explored. ChangeDiff~\cite{zang2024changediff} uses a two-stage diffusion process: a text-to-layout model generates layouts via multi-class prompts, which a layout-to-image model converts into images. It yields coherent, diverse outputs, but its narrow vocabulary limits use in complex scenes. Similarly, Changen2~\cite{zheng2024changen2} simulates semantic change events in a scene's mask and then employs a diffusion transformer to generate the post-event image. Self-supervised training with SAM-extracted contours enables robust zero-shot performance, though its coarse masks limit fine detail. Tang et al.~\cite{tang2024crs} proposed CRS-Diff, a controllable remote sensing generative model that leverages diffusion models with multi-condition guidance. CRS-Diff supports text, metadata, and image conditions, such as sketch, segmentation mask, HED, and road maps, through a new conditional control mechanism that fuses multi-scale features, achieving precise and realistic remote sensing image synthesis. Additionally, Espinosa and Crowley~\cite{espinosa2023generate} propose a ControlNet-based method to synthesize satellite images from OSM maps. Their dataset primarily features green vegetation, and reliance on fixed zoom-level maps without additional temporal or semantic metadata limits its adaptability. Finally, DiffusionSat~\cite{khanna2023diffusionsat} integrates numerical metadata into latent diffusion models for satellite imagery, yielding high-resolution, temporally diverse, and contextually accurate images; however, it fails for prompts with multiple entity classes.

\subsection{Image Editing}
The ultimate goal of this project is to generate a change dataset. For the effective execution of this task, merely generating satellite images is insufficient. This is because, in a change dataset, the "before" and "after" images must be highly correlated in unmodified areas, and a proper change generation method should preserve this dependency. This challenge is quite similar to the classic problem of image editing, where edits must be applied in a manner that maintains the characteristics of the original image.
A significant challenge in this task is the scarcity of training data for "before" and "after" edits, which has spurred the development of numerous training-free approaches. These methods can be broadly categorized into two main groups: (1) methods utilizing cross-attention and (2) methods based on inversions.
Cross-attention-based methods gained popularity with the introduction of the Prompt-to-Prompt paper \cite{hertz2022prompt}. The core idea is that during text-based image generation, the cross-attention between text tokens and latent features provides a powerful mechanism that can be altered to generate edits. A drawback of these methods is their reliance on the internal architecture of the diffusion model, which can limit their applicability. Furthermore, these methods can only edit images generated by the same model and cannot be applied to arbitrary images.

The other approach for training-free editing employs inversion methods. SDEdit \cite{meng2021sdedit} is one of the pioneers in this area. Here, starting from a condition (such as a rough sketch), noise is added to the image, but sparingly, so that some information about the original image is retained. This noisy image is then denoised to produce the edited image. A similar approach is adopted by the img2img functionality of Stable Diffusion \cite{stabilityai2022stable}, which also incorporates support for text conditioning. A major drawback of this approach is that selecting the optimal amount of added noise to balance the trade-off between faithfulness and realism (or diversity) is quite challenging.
One way to address this problem is by using inversions based on a deterministic method like DDIM \cite{song2020denoising}. In DDIM inversion, after the noise addition phase, if conditions (such as the caption) remain unchanged, the denoised image is guaranteed to be identical to the initial image. This offers the strong fidelity that was lacking in previous methods. However, DDIM inversion relies on the assumption that noise at steps $t$ and $t+1$ are very close. This assumption can lead to inaccuracies, which are amplified when using classifier-free guidance. More recent works, such as null-text inversion \cite{mokady2023null} and DDPM inversion \cite{huberman2024edit} have been proposed to address this issue.

%% file: neurips_2025.bbl
\begin{thebibliography}{10}

\bibitem{OSMSlippyMap}
Slippy map tilenames.
\newblock OpenStreetMap Wiki, 2025.
\newblock Accessed May 9, 2025.

\bibitem{stabilityai2022stable}
Stability AI.
\newblock Stable diffusion v2.
\newblock \url{https://stability.ai/blog/stable-diffusion-v2-release}, 2022.

\bibitem{bastani2021updating}
Favyen Bastani, Songtao He, Satvat Jagwani, Mohammad Alizadeh, Hari Balakrishnan, Sanjay Chawla, Sam Madden, and Mohammad~Amin Sadeghi.
\newblock Updating street maps using changes detected in satellite imagery.
\newblock In {\em Proceedings of the 29th International Conference on Advances in Geographic Information Systems}, pages 53--56, 2021.

\bibitem{cheng2023layoutdiffuse}
Jiaxin Cheng, Xiao Liang, Xingjian Shi, Tong He, Tianjun Xiao, and Mu~Li.
\newblock Layoutdiffuse: Adapting foundational diffusion models for layout-to-image generation.
\newblock {\em arXiv preprint arXiv:2302.08908}, 2023.

\bibitem{christie2018functional}
Gordon Christie, Neil Fendley, James Wilson, and Ryan Mukherjee.
\newblock Functional map of the world.
\newblock In {\em Proceedings of the IEEE Conference on Computer Vision and Pattern Recognition}, pages 6172--6180, 2018.

\bibitem{dhariwal2021diffusion}
Prafulla Dhariwal and Alexander Nichol.
\newblock Diffusion models beat {GAN}s on image synthesis.
\newblock {\em Advances in Neural Information Processing Systems}, 34:8780--8794, 2021.

\bibitem{espinosa2023generate}
Miguel Espinosa and Elliot~J Crowley.
\newblock Generate your own scotland: Satellite image generation conditioned on maps.
\newblock {\em arXiv preprint arXiv:2308.16648}, 2023.

\bibitem{goodfellow2014generative}
Ian Goodfellow, Jean Pouget-Abadie, Mehdi Mirza, Bing Xu, David Warde-Farley, Sherjil Ozair, Aaron Courville, and Yoshua Bengio.
\newblock Generative adversarial nets.
\newblock {\em Advances in Neural Information Processing Systems}, 27, 2014.

\bibitem{he2021context}
Sen He, Wentong Liao, Michael~Ying Yang, Yongxin Yang, Yi-Zhe Song, Bodo Rosenhahn, and Tao Xiang.
\newblock Context-aware layout to image generation with enhanced object appearance.
\newblock In {\em Proceedings of the IEEE/CVF conference on computer vision and pattern recognition}, pages 15049--15058, 2021.

\bibitem{hertz2022prompt}
Amir Hertz, Ron Mokady, Jay Tenenbaum, Kfir Aberman, Yael Pritch, and Daniel Cohen-Or.
\newblock Prompt-to-prompt image editing with cross attention control.(2022).
\newblock {\em URL https://arxiv. org/abs/2208.01626}, 1, 2022.

\bibitem{ho2020denoising}
Jonathan Ho, Ajay Jain, and Pieter Abbeel.
\newblock Denoising diffusion probabilistic models.
\newblock {\em Advances in Neural Information Processing Systems}, 33:6840--6851, 2020.

\bibitem{huberman2024edit}
Inbar Huberman-Spiegelglas, Vladimir Kulikov, and Tomer Michaeli.
\newblock An edit friendly ddpm noise space: Inversion and manipulations.
\newblock In {\em Proceedings of the IEEE/CVF Conference on Computer Vision and Pattern Recognition}, pages 12469--12478, 2024.

\bibitem{isola2017image}
Phillip Isola, Jun-Yan Zhu, Tinghui Zhou, and Alexei~A Efros.
\newblock Image-to-image translation with conditional adversarial networks.
\newblock In {\em Proceedings of the IEEE conference on computer vision and pattern recognition}, pages 1125--1134, 2017.

\bibitem{kazemi2019time2vec}
Seyed~Mehran Kazemi, Rishab Goel, Sepehr Eghbali, Janahan Ramanan, Jaspreet Sahota, Sanjay Thakur, Stella Wu, Cathal Smyth, Pascal Poupart, and Marcus Brubaker.
\newblock Time2vec: Learning a vector representation of time.
\newblock {\em arXiv preprint arXiv:1907.05321}, 2019.

\bibitem{khanna2023diffusionsat}
Samar Khanna, Patrick Liu, Linqi Zhou, Chenlin Meng, Robin Rombach, Marshall Burke, David Lobell, and Stefano Ermon.
\newblock Diffusionsat: A generative foundation model for satellite imagery.
\newblock {\em arXiv preprint arXiv:2312.03606}, 2023.

\bibitem{kingma2013auto}
Diederik~P Kingma and Max Welling.
\newblock Auto-encoding variational bayes.
\newblock {\em arXiv preprint arXiv:1312.6114}, 2013.

\bibitem{klemmer2025satclip}
Konstantin Klemmer, Esther Rolf, Caleb Robinson, Lester Mackey, and Marc Rußwurm.
\newblock Satclip: Global, general-purpose location embeddings with satellite imagery.
\newblock {\em Proceedings of the AAAI Conference on Artificial Intelligence}, 39(4):4347--4355, Apr. 2025.

\bibitem{lee2015geospatial}
Jae-Gil Lee and Minseo Kang.
\newblock Geospatial big data: challenges and opportunities.
\newblock {\em Big Data Research}, 2(2):74--81, 2015.

\bibitem{10887344}
Yaxian Lei, Xiaochong Tong, Chunping Qiu, Haoshuai Song, Congzhou Guo, and He~Li.
\newblock Spatial-aware remote sensing image generation from spatial relationship descriptions.
\newblock {\em IEEE Geoscience and Remote Sensing Letters}, 22:1--5, 2025.

\bibitem{li2022review}
Jiayi Li, Xin Huang, Lilin Tu, Tao Zhang, and Leiguang Wang.
\newblock A review of building detection from very high resolution optical remote sensing images.
\newblock {\em GIScience \& Remote Sensing}, 59(1):1199--1225, 2022.

\bibitem{meng2021sdedit}
Chenlin Meng, Yutong He, Yang Song, Jiaming Song, Jiajun Wu, Jun-Yan Zhu, and Stefano Ermon.
\newblock Sdedit: Guided image synthesis and editing with stochastic differential equations.
\newblock {\em arXiv preprint arXiv:2108.01073}, 2021.

\bibitem{mokady2023null}
Ron Mokady, Amir Hertz, Kfir Aberman, Yael Pritch, and Daniel Cohen-Or.
\newblock Null-text inversion for editing real images using guided diffusion models.
\newblock In {\em Proceedings of the IEEE/CVF conference on computer vision and pattern recognition}, pages 6038--6047, 2023.

\bibitem{OpenStreetMap}
{OpenStreetMap contributors}.
\newblock {Planet dump retrieved from https://planet.osm.org }.
\newblock \url{ https://www.openstreetmap.org }, 2017.

\bibitem{qu2023layoutllm}
Leigang Qu, Shengqiong Wu, Hao Fei, Liqiang Nie, and Tat-Seng Chua.
\newblock Layoutllm-t2i: Eliciting layout guidance from llm for text-to-image generation.
\newblock In {\em Proceedings of the 31st ACM International Conference on Multimedia}, pages 643--654, 2023.

\bibitem{radford2021learning}
Alec Radford, Jong~Wook Kim, Chris Hallacy, Aditya Ramesh, Gabriel Goh, Sandhini Agarwal, Girish Sastry, Amanda Askell, Pamela Mishkin, Jack Clark, Gretchen Krueger, and Ilya Sutskever.
\newblock Learning transferable visual models from natural language supervision.
\newblock In {\em Proceedings of the 38th International Conference on Machine Learning}, 2021.

\bibitem{reynard2018five}
Darcy Reynard.
\newblock Five classes of geospatial data and the barriers to using them.
\newblock {\em Geography compass}, 12(4):e12364, 2018.

\bibitem{rombach2022high}
Robin Rombach, Andreas Blattmann, Dominik Lorenz, Patrick Esser, and Bj{\"o}rn Ommer.
\newblock High-resolution image synthesis with latent diffusion models.
\newblock In {\em Proceedings of the IEEE/CVF conference on computer vision and pattern recognition}, pages 10684--10695, 2022.

\bibitem{date2vec}
Surya~Kant Sahu.
\newblock Date2vec: Pretrained embeddings for date-time.
\newblock \url{https://github.com/ojus1/Date2Vec}, 2021.
\newblock Accessed: 2025-09-01.

\bibitem{sohl2015deep}
Jascha Sohl-Dickstein, Eric Weiss, Niru Maheswaranathan, and Surya Ganguli.
\newblock Deep unsupervised learning using nonequilibrium thermodynamics.
\newblock In {\em International Conference on Machine Learning}, pages 2256--2265. PMLR, 2015.

\bibitem{song2020denoising}
Jiaming Song, Chenlin Meng, and Stefano Ermon.
\newblock Denoising diffusion implicit models.
\newblock {\em arXiv preprint arXiv:2010.02502}, 2020.

\bibitem{tang2024crs}
Datao Tang, Xiangyong Cao, Xingsong Hou, Zhongyuan Jiang, Junmin Liu, and Deyu Meng.
\newblock Crs-diff: Controllable remote sensing image generation with diffusion model.
\newblock {\em IEEE Transactions on Geoscience and Remote Sensing}, 2024.

\bibitem{zang2024changediff}
Qi~Zang, Jiayi Yang, Shuang Wang, Dong Zhao, Wenjun Yi, and Zhun Zhong.
\newblock Changediff: A multi-temporal change detection data generator with flexible text prompts via diffusion model.
\newblock {\em arXiv preprint arXiv:2412.15541}, 2024.

\bibitem{zang2025changediff}
Qi~Zang, Jiayi Yang, Shuang Wang, Dong Zhao, Wenjun Yi, and Zhun Zhong.
\newblock Changediff: A multi-temporal change detection data generator with flexible text prompts via diffusion model.
\newblock In {\em Proceedings of the AAAI Conference on Artificial Intelligence}, volume~39, pages 9763--9771, 2025.

\bibitem{zhang2023adding}
Lvmin Zhang, Anyi Rao, and Maneesh Agrawala.
\newblock Adding conditional control to text-to-image diffusion models.
\newblock In {\em Proceedings of the IEEE/CVF international conference on computer vision}, pages 3836--3847, 2023.

\bibitem{zhang2024cc}
Mu~Zhang, Yunfan Liu, Yue Liu, Yuzhong Zhao, and Qixiang Ye.
\newblock Cc-diff: enhancing contextual coherence in remote sensing image synthesis.
\newblock {\em arXiv preprint arXiv:2412.08464}, 2024.

\bibitem{zhao2020layout2image}
Bo~Zhao, Weidong Yin, Lili Meng, and Leonid Sigal.
\newblock Layout2image: Image generation from layout.
\newblock {\em International journal of computer vision}, 128(10):2418--2435, 2020.

\bibitem{zheng2023layoutdiffusion}
Guangcong Zheng, Xianpan Zhou, Xuewei Li, Zhongang Qi, Ying Shan, and Xi~Li.
\newblock Layoutdiffusion: Controllable diffusion model for layout-to-image generation.
\newblock In {\em Proceedings of the IEEE/CVF Conference on Computer Vision and Pattern Recognition}, pages 22490--22499, 2023.

\bibitem{zheng2024changen2}
Zhuo Zheng, Stefano Ermon, Dongjun Kim, Liangpei Zhang, and Yanfei Zhong.
\newblock Changen2: Multi-temporal remote sensing generative change foundation model.
\newblock {\em IEEE Transactions on Pattern Analysis and Machine Intelligence}, 2024.

\bibitem{zhu2017unpaired}
Jun-Yan Zhu, Taesung Park, Phillip Isola, and Alexei~A Efros.
\newblock Unpaired image-to-image translation using cycle-consistent adversarial networks.
\newblock In {\em Proceedings of the IEEE international conference on computer vision}, pages 2223--2232, 2017.

\bibitem{zhu2017deep}
Xiao~Xiang Zhu, Devis Tuia, Lichao Mou, Gui-Song Xia, Liangpei Zhang, Feng Xu, and Friedrich Fraundorfer.
\newblock Deep learning in remote sensing: A comprehensive review and list of resources.
\newblock {\em IEEE geoscience and remote sensing magazine}, 5(4):8--36, 2017.

\end{thebibliography}
